\documentclass[sigconf]{acmart}

\AtBeginDocument{%
  \providecommand\BibTeX{{%
    \normalfont B\kern-0.5em{\scshape i\kern-0.25em b}\kern-0.8em\TeX}}}

\setcopyright{none}
\renewcommand\footnotetextcopyrightpermission[1]{}
\thanks{The paper is accepted for presentation at 2022 IEEE/RSJ International Conference on Intelligent Robots and Systems (IROS).

This work was supported by the US National Science Foundation award \#00074041, a UMII-MnDRIVE Fellowship, and the MnRI Seed Grant. The authors are with the Department of Computer Science \& Engineering and the Minnesota Robotics Institute, University of Minnesota, MN, USA. \tt\small \{$^{1}$enan0001, $^{2}$fulto081, $^{3}$junaed\}@umn.edu}%

\newcommand{\eg}{\emph{e.g.}, }
\newcommand{\ie}{\emph{i.e.}, } 
\newcommand{\etal}{\emph{et al.} } 
\usepackage[inline]{enumitem}
\usepackage{subcaption}
\usepackage{graphicx}
\usepackage{multirow}
\usepackage{caption}
\usepackage{hyperref}
\usepackage{diagbox}
\usepackage{textcomp}

\begin{document}

\title{Robotic Detection of a Human-Comprehensible Gestural Language for Underwater Multi-Human-Robot Collaboration}

\author{Sadman Sakib Enan$^1$, Michael Fulton$^{2}$ and Junaed Sattar$^{3}$}

\begin{abstract}
In this paper, we present a motion-based robotic communication framework that enables non-verbal communication among autonomous underwater vehicles (AUVs) and human divers.
We design a gestural language for AUV-to-AUV communication which can be easily understood by divers observing the conversation unlike typical radio frequency, light, or audio based AUV communication.
To allow AUVs to visually understand a gesture from another AUV, we propose a deep network (RRCommNet) which exploits a self-attention mechanism to learn to recognize each message by extracting maximally discriminative spatio-temporal features.
We train this network on diverse simulated and real-world data.
Our experimental evaluations, both in simulation and in closed-water robot trials, demonstrate that the proposed RRCommNet architecture is able to decipher gesture-based messages with an average accuracy of 88-94\% on simulated data, 73-83\% on real data (depending on the version of the model used).
Further, by performing a message transcription study with human participants, we also show that the proposed language can be understood by humans, with an overall transcription accuracy of 88\%.
Finally, we discuss the inference runtime of RRCommNet on embedded GPU hardware, for real-time use on board AUVs in the field.
\end{abstract}

\maketitle
\thispagestyle{empty}
\pagestyle{empty}

\section{Introduction}
\begin{figure}[t]
    \centering
    \vspace{2mm}
    \includegraphics[width=.99\linewidth]{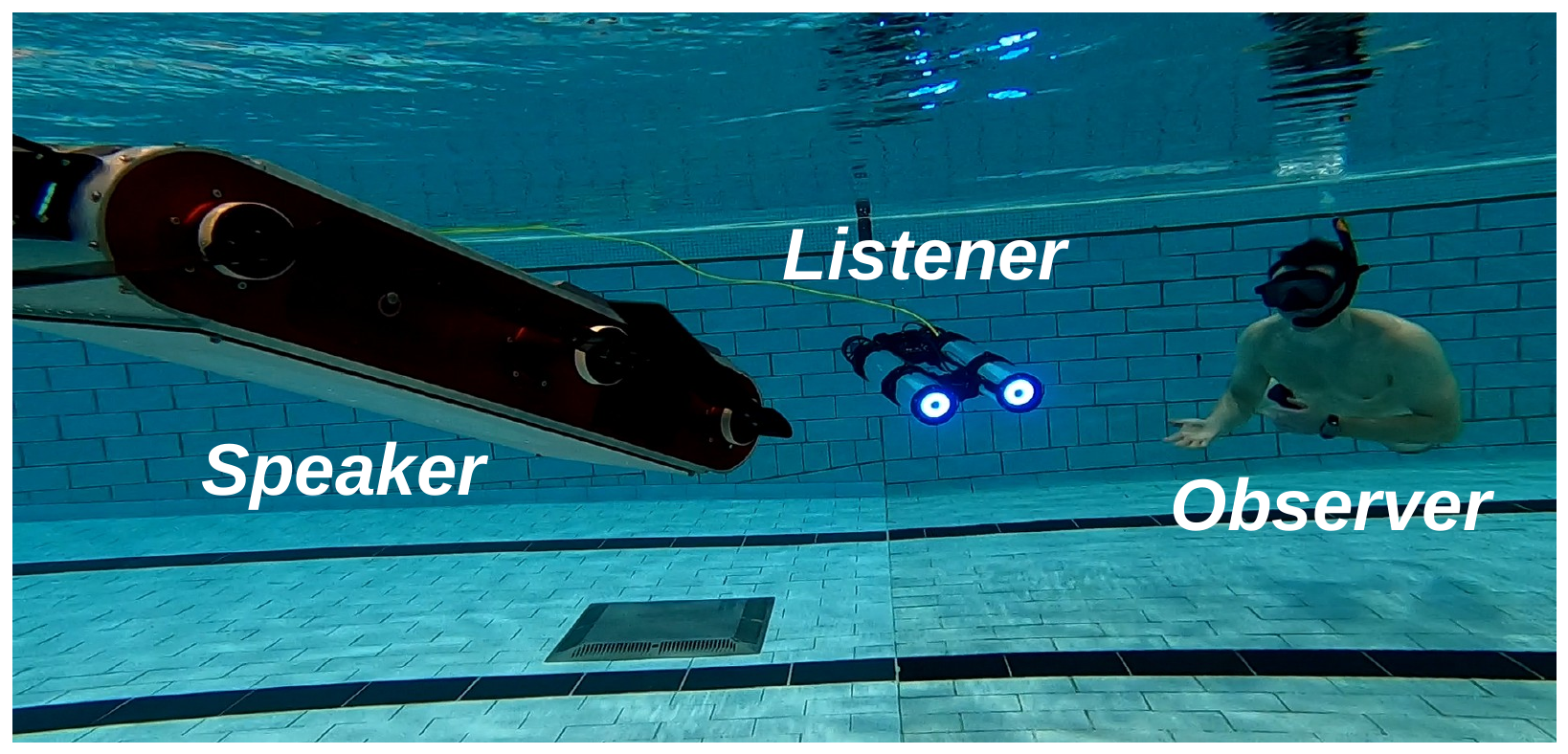}
    \caption{An underwater gestural communication framework where the speaker robot is communicating back with the listener robot by making a \textit{nodding motion} to mean YES which a human observer is also able to understand.}
    \label{fig:intro}
    \vspace{-4mm}
\end{figure}

Over the last several decades, applications of autonomous underwater vehicles (AUVs)~\cite{sattar2008enabling,edge2020design} have multiplied and diversified (\eg environmental monitoring and mapping~\cite{fulton2019robotic,weidner2017underwater}, submarine cables and wreckage inspection~\cite{bingham2010robotic}, search and navigation~\cite{koreitem2020one-shot,xanthidis2020navigation}), driven by ever-increasing on-board computational power, increased affordability, and ease of use. 
The majority of these applications involve multiple AUVs and/or their human diver companions, often interacting with one another to work effectively as a team~\cite{islam2018understanding,hong2020visual}. 
Thus, robust underwater human-to-robot and robot-to-human interaction capabilities are of utmost value.
A common language comprehensible to both humans and other AUVs would greatly enhance such underwater multi-human-robot (m/HRI) missions (see Fig.~\ref{fig:intro}). 

When designing such a communication protocol, challenges unique to the underwater domain need to be considered. 
Traditional sensory mediums, such as radio and other electromagnetic (EM) modalities, suffer from signal attenuation and degradation~\cite{qureshi2016rf} underwater which limit their use to mostly surface operations. Although acoustic signals work quite well in underwater settings~\cite{farr2010an}, these types of inter-AUV communication signals are typically incomprehensible to humans.
Our recent work utilizing robot motion for AUV-to-diver communication has demonstrated that motion can be used to communicate with divers~\cite{fulton2019robot,fulton_rcvm-thri_2022}.
Similarly, research on the use of motion for inter-AUV communication has shown the same for AUV-to-AUV communication~\cite{koreitem2019underwater}.
However, the two capabilities have yet to be combined in a \textbf{single language for multi-human-robot communication}.
The natural choice for AUV perception of gestures is the use of vision, as many AUVs are equipped with low-cost vision sensors.
While vision underwater can be impacted by water quality and turbidity, promising results achieved in the improvement of underwater vision (\eg\cite{islam2020fast,islam2020underwater}) provide increased robustness in AUV visual perception.
\begin{figure*}[t]
    \centering
    \vspace{2mm}
    \includegraphics[width=0.99\linewidth]{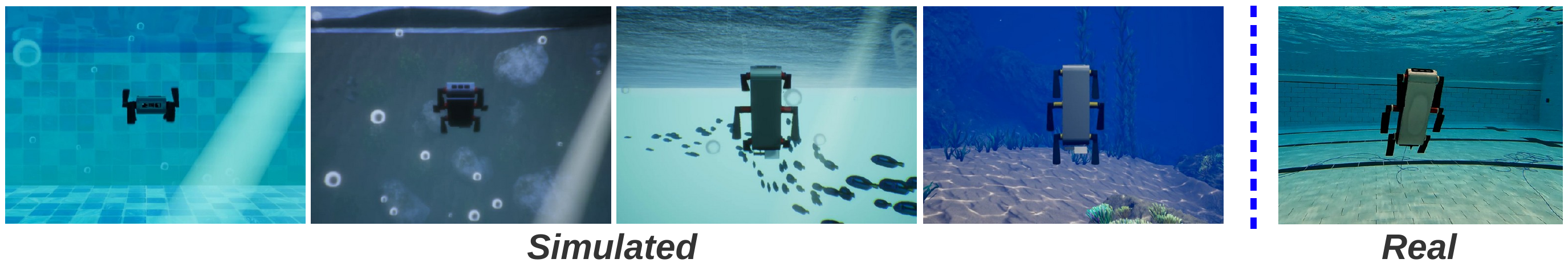}
    \caption{The Aqua robot is performing ASCEND message in both simulated and real underwater environments.}
    \label{fig:sample}
    \vspace{-4mm}
\end{figure*}

To create a natural, comprehensible, and accurate language for robots in an m/HRI context, we propose to use robot motion to design \textit{gestural messages} as in~\cite{fulton2019robot, fulton_rcvm-thri_2022}, for both robot-to-human and robot-to-robot communications (see Fig.~\ref{fig:sample}). 
We design these messages in simulated underwater environments~\cite{games2004unreal}, using computer-aided design (CAD) renderings of a six-legged AUV named Aqua~\cite{dudek2007aqua}.
Additionally, we implement the gestural messages on board an actual Aqua robot using the Robot Operating System (ROS)~\cite{quigley2009ros}.
For a robot to interpret these messages, we propose a recognition network, \underline{R}obot-to-\underline{R}obot \underline{Comm}unication \underline{Net}work (\textbf{RRCommNet}), which learns salient spatio-temporal features from the gestural messages using a self-attention mechanism. 
After training RRCommNet on simulated and real-world data, our experiments show the recognition accuracy of RRCommNet to be approximately $94$\% on simulated data and $83$\% on real data.
Our experiments are undertaken on simulated and closed water (pool) environments, as training and evaluation in these environments is a necessary prerequisite to the creation and evaluation of a system in the field.
Based on our previous experience with computer vision-based methods in underwater field environments (\eg \cite{islam2020fast,islam2018dynamic}), we are aware of the challenge the transition from simulated/pool to field environments poses, but are prepared for those challenges.
We also show that we can improve the inference time of RRCommNet while compromising the accuracy somewhat (simulated: $88$\%, real: $73$\%) by down-sampling the input video by half.
Finally, through a transcription experiment, we show that humans can comprehend a conversation between two AUVs using the gestural messages with a transcription accuracy of $88$\%.
Together, these two evaluations demonstrate that our gestural communication system can be used for accurate communication by an AUV in an m/HRI context. Thus, in this paper, we:
\begin{enumerate}
    \item Propose a gestural language for AUV-to-AUV communication,
    \item Create an end-to-end gestural message recognition network, RRCommNet, to interpret underwater communication between AUVs,
    \item Perform experiments in both simulated and real underwater environments to validate the performance of the proposed gestural message recognition network, and
    \item Conduct a study which demonstrates that the proposed gestural language can be understood by humans.
\end{enumerate}
\section{Related Work}
Underwater communication has largely focused on \textit{human-to-robot communication}, \ie regulating underwater robots based on human inputs. 
One common approach is to control the robot via high-speed tethered communication~\cite{aoki1997development}, however, direct communications are often preferred in missions as opposed to using tethered connections. 
A number of direct communication techniques do not require additional robot hardware; \eg using fiducial markers~\cite{sattar2007fourier} or hand gestures~\cite{islam2018dynamic,chavez_2021_gesture}.
For \textit{robot-to-human communication}, on the other hand, the use of small displays is by far the most popular method~\cite{miskovic_caddycognitive_2016} although it has significant limitations in readability at distance, at an angle, and under low water quality. 
There are alternatives, such as the use of a bi-directional communication device~\cite{verzijlenberg2010swimming} and the use of light to represent simple ideas~\cite{demarco2014underwater}; however, they either require the addition of dedicated communication devices or lack functionalities. 
The use of robot motion~\cite{fulton2019robot} for information or \textit{affective} displays~\cite{bethel2007survey} for appearance-constrained robots has seen encouraging results in communicating with humans. 

In contrast to human-in-the-loop communication, underwater \textit{robot-to-robot communication} systems are significantly less studied, primarily due to the fact that robots are not as perceptive as humans and therefore need sophisticated algorithms to understand what other robots are trying to communicate. In~\cite{dudek1995experiments}, body markings (helical drawings) are used by AUVs to communicate relative pose information. 
More recently, Koreitem \etal\cite{koreitem2019underwater} propose a communication system for underwater robots where the communication messages are represented using full-body gestures and optimal variable-length prefix codes. However, this technique involves multiple steps to devise different messages and the proposed CNN learns indirectly from them. As a result, adding additional messages is not straightforward. 
In this work, we focus on methods such as activity recognition that can directly learn from gesture-based messages.

\begin{table*}[t]
\centering
\vspace{2mm}
\caption{Listing of communication messages [D=Directional, I/C=Information/Command, CC=Conversation Control].} 
\begin{tabular}{l|c|c|c}
  \hline
  \begin{tabular}{@{}c@{}}\textbf{Communication} \\ \textbf{Message}\end{tabular} & \textbf{Description and Type}  & \textbf{Definition} & \begin{tabular}{@{}c@{}}\textbf{Average} \\ \textbf{Duration} \\\textbf{ (secs)}\end{tabular}\\ \hline \hline
  BATTERY LOW &  Signal low battery (I/C)  & Roll $360^{\circ}$ twice while moving
down vertically & $4.28$\\
  \hline
  \begin{tabular}{@{}l@{}}START \\COMMUNICATION\end{tabular}  &  Begin robot-to-robot communication (CC)  & Roll, pitch, and yaw $45^\circ$ at the same time twice & $16.71$\\ 
  \hline
  ASCEND &  Go up to a certain
depth (D) &  Pitch up $90^{\circ}$ vertically and go up & $3.18$ \\ 
  \hline
  DESCEND &  Go down to a certain
depth (D) &  Pitch down $90^{\circ}$ vertically and go down & $ 3.28$\\
  \hline
  FOLLOW ME & Instruct another robot to follow it (D) & \begin{tabular}{@{}l@{}}Pitch and yaw $35^\circ$ to the right, comeback to original \\  position; do this twice and then roll $45^\circ$, make a $180^\circ$ \\ turn while unrolling and finally, go forward \end{tabular} & $6.93$ \\ 
  \hline
  DANGER &  Danger nearby (I/C) & \begin{tabular}{@{}l@{}}Yaw left and right thrice; then roll $45^\circ$, make a $180^\circ$ \\ turn while unrolling \end{tabular}  & $7.44$\\
  \hline
  COLLECT DATA &  Start data collection (I/C) & \begin{tabular}{@{}l@{}}Pitch and yaw $45^\circ$ to the right, comeback to original \\  position, do the same to the left; perform this twice \end{tabular} & $8.32$ \\ 
  \hline
  START MAPPING & Map the environment (I/C) &  Yaw $360^{\circ}$ while pitching up and down & $ 4.32$ \\ 
  \hline
  GO TO LOCATION &  Go to a specific location (D) & \begin{tabular}{@{}l@{}} Roll $45^\circ$, move forward and backward twice, then go \\to a location\end{tabular}  & $7.48$\\ 
  \hline
  U-TURN &  Danger nearby, make a u-turn (D)  &  Roll $80^{\circ}$ and make a $180^{\circ}$ turn while unrolling & $ 4.22$ \\
  \hline
  HELP &  Call for help (I/C) & Roll $45^\circ$, circle in a small loop twice & $19.95$\\ 
  \hline
  \begin{tabular}{@{}l@{}}EMERGENCY\\SURFACING\end{tabular}  &  Go to the surface of
the water (D) &  Pitch up $90^{\circ}$ and roll $360^{\circ}$ twice  & $ 3.46$ \\ 
  \hline
  STOP &  Stop doing whatever
you’re doing (CC) &  Yaw $360^{\circ}$ twice before stopping & $ 6.52$ \\
  \hline
  NO &  Disagreement (CC) &  Yaw left and right twice & $ 4.31$ \\ 
  \hline
  YES &  Agreement (CC)  &  Pitch down and up twice & $ 4.26$ \\
  \hline
\end{tabular}
\label{tab:msg_des}
\vspace{-2mm}
\end{table*}

Activity recognition (AR) is a well-studied problem in computer vision and robotics, with research spanning over two decades~\cite{wang2018non,ji20133d,wang2016temporal}.
The goal is to predict an activity class from a large pool of human activities involving exercise, sport, instrument performance, everyday life, etc. 
The general solution to this problem is to learn robust spatio-temporal features from different activity classes which are contained in small video clips. 
A variety of methods have been used to integrate the temporal information with the spatial, such as pooling~\cite{girdhar2017actionvlad}, fusion~\cite{karpathy2014large-scale}, recurrent~\cite{li2018videolstm}, two-stream~\cite{simonyan2014two-stream,feichtenhofer2019slowfast}, and $3$D architectures~\cite{carreira2017quo}. 
These techniques perform well mainly because of their ability to learn from publicly available large datasets involving human activities. 
The lack of such datasets is the primary contributing factor behind the absence of activity recognition techniques for robot actions in the literature. 
Therefore, robotic researchers often use synthetic data to validate their proposed methods before fine-tuning them for real robotic platforms~\cite{koreitem2018synthetically,demelo2020vision-based}. 
Furthermore, unlike traditional human activity recognition datasets, the communication messages we have designed for underwater robots have high similarities in terms of background activities and the overall appearance of the robot motions. 
Additionally, there are cases in which a portion of a gesture's motion can be common to multiple gestures, leading to a strong resemblance between two gestures when considering small portions of their motion.
As a result, AR algorithms that perform well for human activity recognition may not work well for underwater robots. 
Recent research on natural language processing suggests that one can effectively learn the underlying sequential relations by using a \textit{self-attention} mechanism~\cite{vaswani2017attention}, which has also shown encouraging results in AR recently~\cite{girdhar2019video, kalfaoglu2020late}. 
Inspired by such work, we believe that highly robust spatio-temporal features can be learned from our gesture-based communication messages using a similar attention mechanism.

\section{GESTURAL LANGUAGE RECOGNITION FRAMEWORK}
\label{sec:rrcomm}
\subsection{Designing the Communication Messages} \label{design_msg}
Since the purpose of using motion-based gestural communication is to enable human understanding of robot-to-robot conversations, we draw inspiration from our previous work on robot-to-human underwater communication methods~\cite{fulton2019robot,fulton_rcvm-thri_2022}, where we have shown that humans are able to identify gestural messages with reasonable accuracy. 
We create a library of communication messages based on our experience working with AUVs in real underwater environments and discussions with potential end-users (\eg marine biologists). Our library includes three types of messages: \begin{enumerate*}
    \item \textit{Directional (D):} which relate to a notification or command of some directional movement, 
    \item \textit{Information/Command (I/C):} which provide information or give direct commands to another robot, and
    \item \textit{Conversation Control (CC):} which are responses to questions or commands.
\end{enumerate*}
Table~\ref{tab:msg_des} shows the complete library of communication messages with their description, type, definition, and average duration. 

\subsection{Implementation Of Gestures}
\begin{figure}[t]
    \centering
    \includegraphics[width=0.99\linewidth]{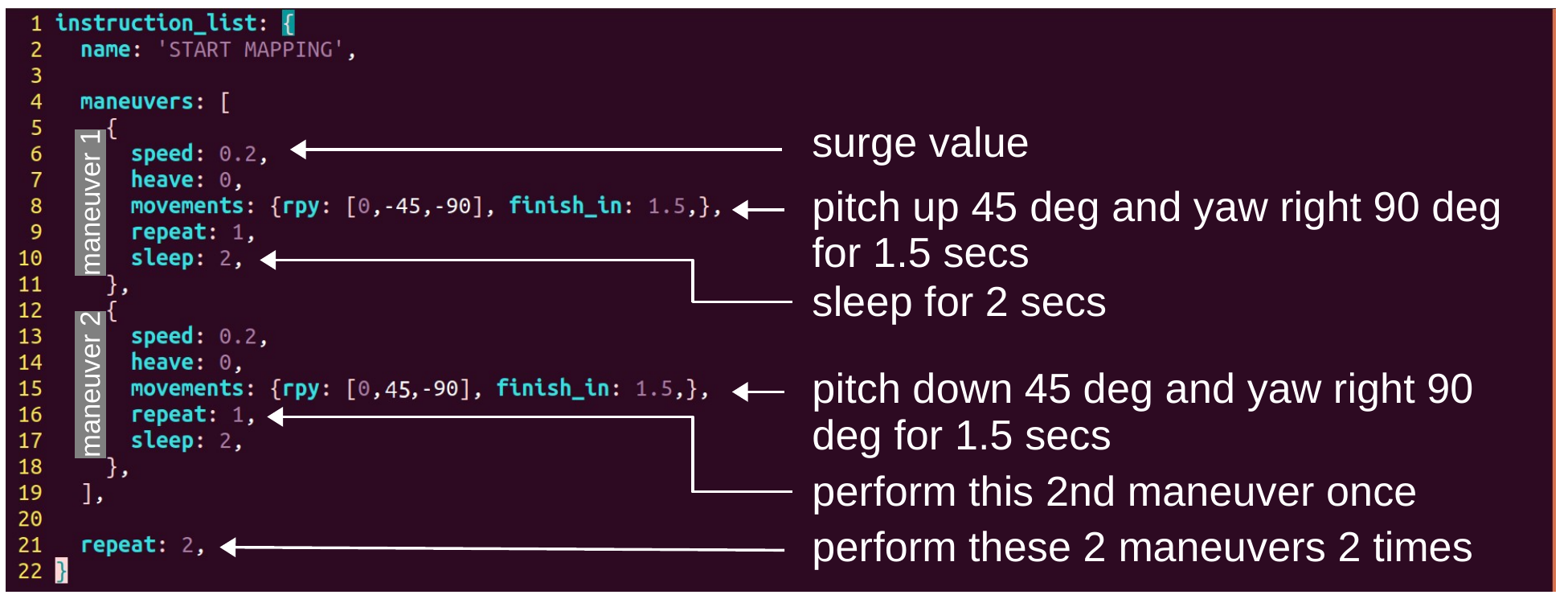}
    \caption{The description of the configuration file for the START MAPPING message.}
    \label{fig:start_mapping}
    \vspace{-4mm}
\end{figure}

We first implement the gestural messages in simulated underwater environments using Unreal Engine visual programming~\cite{games2004unreal}, with CAD renderings of the Aqua robot. 
We use three self-made environments: pool, lake, ocean, and a separate ready-made ocean environment~\cite{oceanpack} as the visual environments the gestures will be performed in. 
To implement the messages in real underwater environments with the Aqua robot, we use the ROS libraries~\cite{quigley2009ros} to create a ROS package named \texttt{rrcomm} which operates as a gestural message generator. 
The generator can produce $15$ communication messages using a user-defined configuration file, an example of which can be seen in Fig.~\ref{fig:start_mapping} for the START MAPPING message. 
The configuration file contains a definition of the gesture as a set of timed motions expressed in roll, pitch, yaw, surge, and heave velocity percentages. Our ROS node parses these configuration files to implement the robot maneuvers using the robot motion controller~\cite{giguere2013wide-speed}.

\subsection{Gestural Message Dataset} \label{sec:data}
To train a neural network capable of recognizing our gestural messages, we first create a dataset by recording the $15$ defined gestural messages (see Table~\ref{tab:msg_des}) as full-color videos in both simulated and real underwater environments. 
To improve the robustness of our simulated training data, we introduce a number of variations within the simulated environment and the robot model, giving rise to $25$ different environmental conditions. 
Specifically, we vary the surface texture, object type, hydrodynamics, water visibility, and color of the robot while keeping the message definitions unchanged. 
For real-world data, we record $5$ instances of each message performed in a closed-water underwater environment. 
For the HELP message, the Aqua robot's PID controller was overshooting the requested target angles, causing the robot to stray away from the defined circular path, which made this implementation invalid for use. In summary, there are a total of $445$ recordings of the $15$ communication messages, of which $375$ are synthetic data and $70$ are real-world data. 
We reserve $73$\% of recordings reserved for training ($300$ synthetic, $28$ real) and the remaining $27$\% for testing ($75$ synthetic, $42$ real).
 
\subsection{Robot-To-Robot Communication Network (\textbf{RRCommNet})}
The RRCommNet network accepts videos of robot gestures as inputs, having dimensions of  
$T \times C \times H \times W$, where $H,W,$ and $C$ are height, width, and color channels of the input, respectively. $T$ is the length of the input chunk in frames. We introduce a skipping mechanism for faster training and inference; if enabled, it uses every other frame from the input chunk (\ie $T/2$ number of frames). This gives rise to two different models, \textit{RRCommNet} and \textit{RRCommNet-Skip}, both of which look at the same temporal context where the latter simply handles a temporally downsampled version of the input. We encode the input using a highly modularized deep neural network architecture, ResNeXt-101~\cite{xie2017aggregated}, which outputs a feature representation of shape ($T' \times C' \times H' \times W'$). We choose ResNeXt-101 as a feature extractor because it can extract highly robust spatio-temporal features by utilizing a \textit{split-transform-aggregation} strategy. With this strategy, the input is split into a few lower-dimensional embeddings, transformed as a group (in a new dimension called the \textit{cardinality} of the architecture), and merged by concatenation. This gives the network the representational power of large and dense layers but at a considerably lower computational complexity.
\begin{figure}[t]
    \centering
    \vspace{2mm}
    \includegraphics[width=0.99\linewidth]{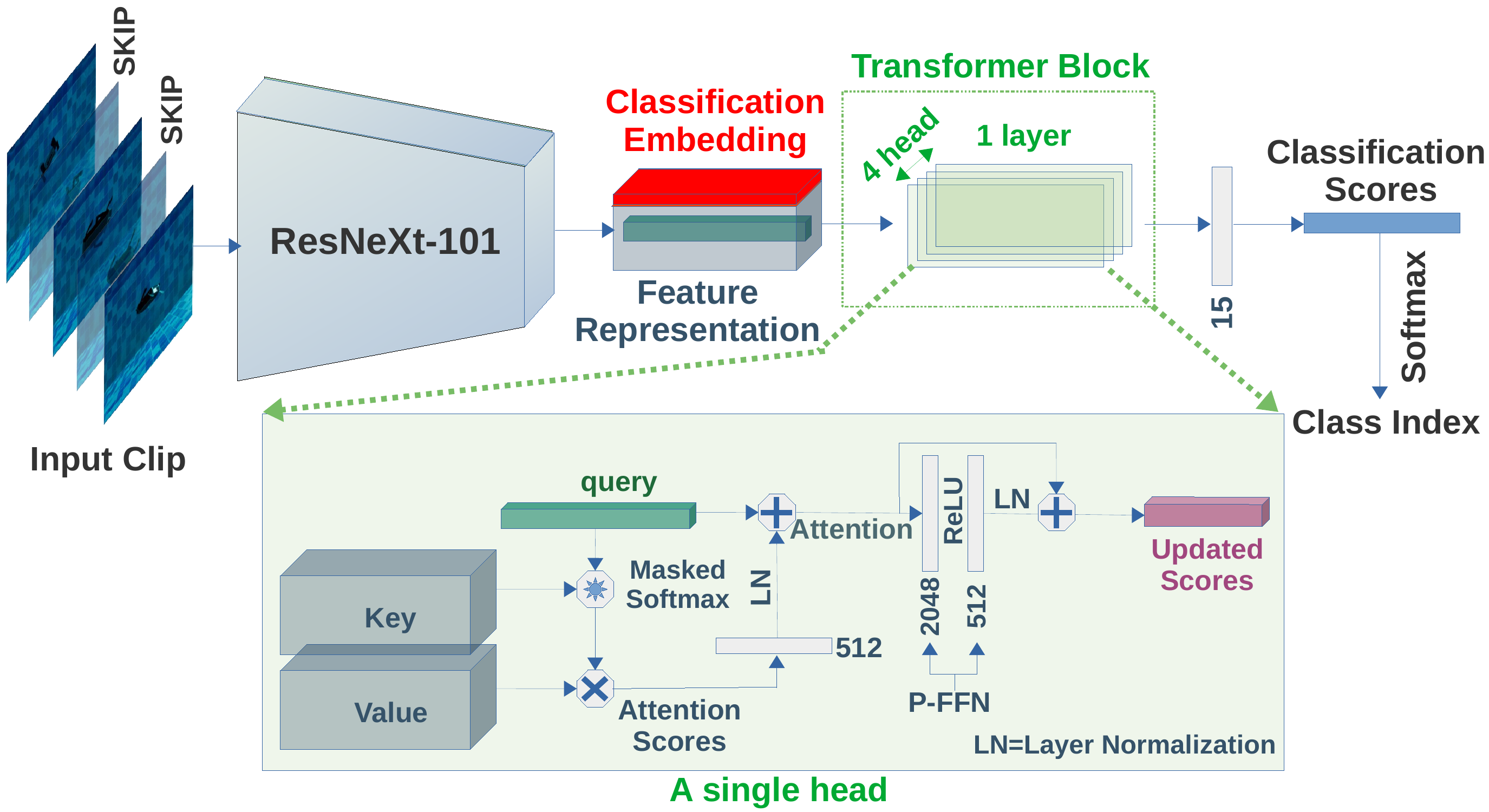}
    \caption{The proposed gestural message recognition framework, RRCommNet. It takes a video clip as input and extracts spatio-temporal features. Self-attention mechanism is performed on the feature representation to get the final classification scores.}
    \label{fig:recog_framework}
    \vspace{-4mm}
\end{figure}

We need to address two issues unique to the gestural messages while designing our recognition framework. First, the gestural messages have high similarities in terms of background
activities and the overall appearance of the robot motions. To address this, we perform average pooling in the spatial dimension while normalizing against the channel dimension on the extracted features to get a narrow but long feature representation having a dimension of ($T'\times C'$). This will give less importance to the background and appearance information. The second and more significant issue is the commonality between portions of pairwise gestures; a strong resemblance can confuse the networks when considering small durations of such pairs of gestures. We address this by using a \textit{self-attention} mechanism (proposed in~\cite{vaswani2017attention}) with bi-directional Transformers (BERT~\cite{devlin2018bert}), as was suggested in~\cite{girdhar2019video,kalfaoglu2020late}. With self-attention, the network will be able to focus on the most salient features while BERT enables temporal information from both directions. Taking these together, the network is able to look into the most salient features within a longer temporal limit enhancing the messages' contexts. With these choices in mind, we design RRCommNet (see Fig.~\ref{fig:recog_framework}) as described below.

First, we take a \textit{query} feature ($Q$) from the extracted feature representation (computed using ResNeXt-101) to compare against all other positions in the feature representation (considered as \textit{key}, $K$). For a gestural message, $Q$ is the small spatial information with a large temporal context. Both query and key are of size $d_{model}(=C')$. The comparison gives rise to a weighted sum of \textit{values} ($V$) which are the self-attention scores. Note that to learn the robust feature representations of $Q$ and $K$, we use linear projection to project them to a lower dimension $d_k$. Specifically, we compute the self-attention as follows:
\begin{align}
   a_{self} &= \frac{QK^T}{\sqrt{d_k}} \nonumber \\
   a_{weights}^i &= \frac{\exp \left(a_{self}(:,:,i)\right)}{\sum_{j=1}^{d_k} \exp \left(a_{self}(:,:,j)\right)}\nonumber \\ 
   A_{self}^i &=  a_{weights}^i V \nonumber
\end{align}
where, $A_{self}$ is the weighted self-attention scores.
\begin{table*}[t]
    \centering
    \vspace{2mm}
    \caption{Results of the human transcription study in recognizing different gestural messages. The values represent \textit{avg. trans. accuracy (\%)$\Large/$avg. confidence (out of 10)}.}
\begin{tabular}{l||cccccccccccccccc}
 \hline
\textbf{Gest. Message}  & BATTERY & START &  &  & FOLLOW &  & COLLECT & START \\
 
 & LOW & COMM. & ASCEND & DESCEND & ME & DANGER & DATA & MAPPING \\
 
\hline
Human Trans. & $94.10/8.30$ & $94.10/7.30$ & $97.10/8.30$ & $94.10/8.30$ & $97.10/8.60$ & $72.10/7.80$ & $85.30/8.30$ & $58.80/7.60$ \\
 
 &  &  &  &  &  &  &  &  \\
 
\hline
\textbf{Gest. Message}  & GO TO &  &  & EMERGENCY &  &  &  &  \\
 
   & LOCATION & U-TURN & HELP & SURFACING & STOP & NO & YES & \textbf{Overall} \\ 
 
 \hline
Human Trans. & $82.40/8.10$ & $87.10/6.40$ & $91.20/8.00$ & $95.60/8.00$ & $73.50/6.40$ & $94.10/8.20$ & $96.30/8.20$ & $88.20/7.90$  \\
\hline

\end{tabular}
\label{tab:human_study}
\vspace{-4mm}
\end{table*}

Note that we use the attention mechanism in a parallel fashion, \ie using multiple attention heads ($h$) and set $d_k=\text{floor}(d_{model}/h)$. The multi-head attention scores are fed through a two layer position-wise feed-forward network (P-FFN) to get the final self-attention scores to represent the salient features from a gestural message. P-FFN is defined as, $\text{P-FFN}(A_{self}) = \max(0, A_{self}W_1 + b_1)W_2 + b_2$, where $W_i$ and $b_i$ are the weights and biases of the respective layers.

Finally, a single linear layer is used for the final classification scores which has exactly $N$ neurons, where $N$ is the total number of communication messages. Note that we perform the final prediction using the classification embedding which we include on top of the feature representation for better classification, as suggested in~\cite{devlin2018bert,kalfaoglu2020late}. Additionally, we include a location embedding to all the locations in the feature representation in order to incorporate positional information in the attention scores. These embeddings are set as learnable parameters. Moreover, while performing the self-attention mechanism, we randomly mask $10\%$ of the feature locations and set their attention scores to zeros which incorporates the bi-directional context learning from BERT.

\subsection{Implementation Details}
We use PyTorch~\cite{paszke2019pytorch} libraries to implement RRCommNet, with input spatial resolution of $320\times 256$ and temporal resolution of $64$ or $32$ (with input skipping). We follow various data augmentation schemes, \eg multi-scale cropping, random horizontal flipping, and normalization, before training the model as described in~\cite{wang2015towards}. For training the overall network, we use similar settings as described in~\cite{kalfaoglu2020late} with the following modifications. We choose batch size of $8$, learning rate of $10^{-4}$ with a scheduler, and ADAMW as the optimizer~\cite{loshchilov2017decoupled}. We use ResNeXt-101 with a \textit{cardinality} of $32$ without the last two layers. The P-FFN uses ReLU non-linearity and has $2$ layers with $2048$ and $512$ neurons, respectively. The dropouts are set to $0.1$. We use one transformer block with four attention heads for the self-attention mechanism. We use $d_{model}=512$ and $d_k=128$. A classification embedding is added on top of the feature representation and a location embedding is added to all the locations in the feature representation in order to incorporate positional information. Both of which are initialized using the normal distribution $\mathcal{N}(0,0.02^2)$. We have trained the network on an Nvidia GeForce RTX 2080 GPU for $200$ epochs with cross-entropy loss and noticed convergence in validation loss, top-1 accuracy, and top-3 accuracy. The network is evaluated on a Intel\textregistered{} Xeon\textregistered{} E5-2650 CPU for realism of inference time results.

\section{HUMAN TRANSCRIPTION STUDY} \label{sec:human_study}
To demonstrate that the proposed gestural language can be effectively understood by humans in an m/HRI joint mission, we undertake a study\footnote{Study (reference no. $00012959$) has been reviewed and approved by the University of Minnesota's Institutional Review Board.} of humans transcribing the conversations of robots using our gestural language, using the QualtricsXM survey platform. 
First, the participants are taught the messages in a random order. 
Then, they are asked to transcribe the conversation shown to them in a simulated video of two robots conversing. A total of $10$ such conversations are shown from one of three different viewpoints: head-on, rotated by $90$ degrees of yaw, and rotated by $90$ degrees of pitch. 
Our population of $34$ participants are randomly assigned to one of these viewpoints, asked to select the displayed gesture's meaning from a drop-down list for each message, and rate their confidence in their answers for each transcription from zero to $10$. Table~\ref{tab:human_study} shows the results of the study. 
The participants are able to correctly transcribe the conversation with an average accuracy of $88.20\%$ and confidence of $7.9$ (out of $10$). 
Here, $\text{avg. accuracy}=\frac{1}{34}\sum_{p=1}^{34}\frac{\text{(correct selections)}_p}{\text{(total shown)}_p}$. 
The participants seem to struggle the most with the messages which RRCommNet performs poorly on (discussed in Sec.~\ref{sec:results}). Closer inspection of such
messages indicates that they include gestures which have
high visual similarity. The results from this study show that humans can comprehend our proposed gestural language, which, along with the results in the following section, fulfills our goals of a gestural language that can be understood by both robots and humans.

\begin{figure}[t]
    \centering
    \vspace{2mm}
    \begin{subfigure}{0.49\linewidth} 
    \centering
        \includegraphics[width=.99\linewidth]{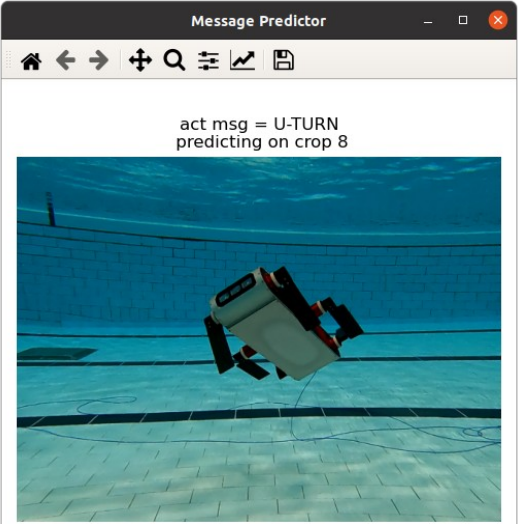}
        \caption{}
    \end{subfigure}
    \begin{subfigure}{0.49\linewidth} 
    \centering
        \includegraphics[width=.99\linewidth]{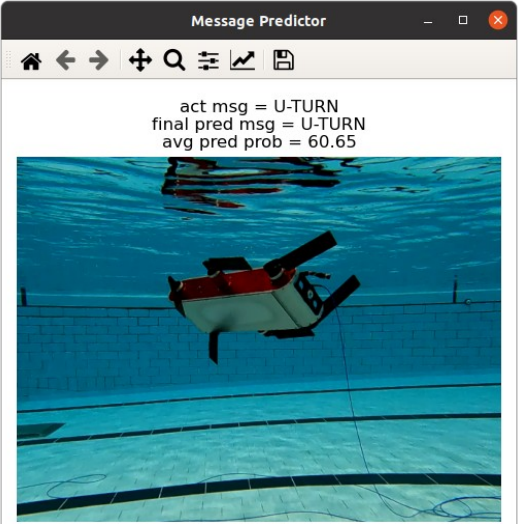}
        \caption{}
    \end{subfigure}
    \caption{Gestural message recognition using RRCommNet. (a) Prediction is being performed on the $8$th crop. (b) Final prediction is made by averaging the recognition probabilities on all $10$ crops.}
    \label{fig:vis}
    \vspace{-4mm}
\end{figure}

\section{EVALUATION OF RRCommNet} \label{sec:eval_rrcomm}
\subsection{Evaluation Procedure and Metrics} \label{sec:eval_process}
The test videos are processed as chunks of either $64$ or $32$ frames (with skipping). We make the final prediction\footnote{In this paper, we use prediction and recognition interchangeably.} on $10$ different cropped versions of the input video clip. We follow the same cropping mechanism as described in~\cite{wang2015towards}, \ie four crops from the corners, one from the center, and the same on a flipped version, for a total of $10$ cropped versions. 
The crops have a spatial dimension of $112\times 112$. 
Therefore, a single RGB test batch is a $5$-dimensional tensor of shape $[10,64,3,112,112]$ or $[10,32,3,112,112]$ (with skipping).
\begin{figure}
    \centering
    \vspace{2mm}
    \includegraphics[width=.99\linewidth]{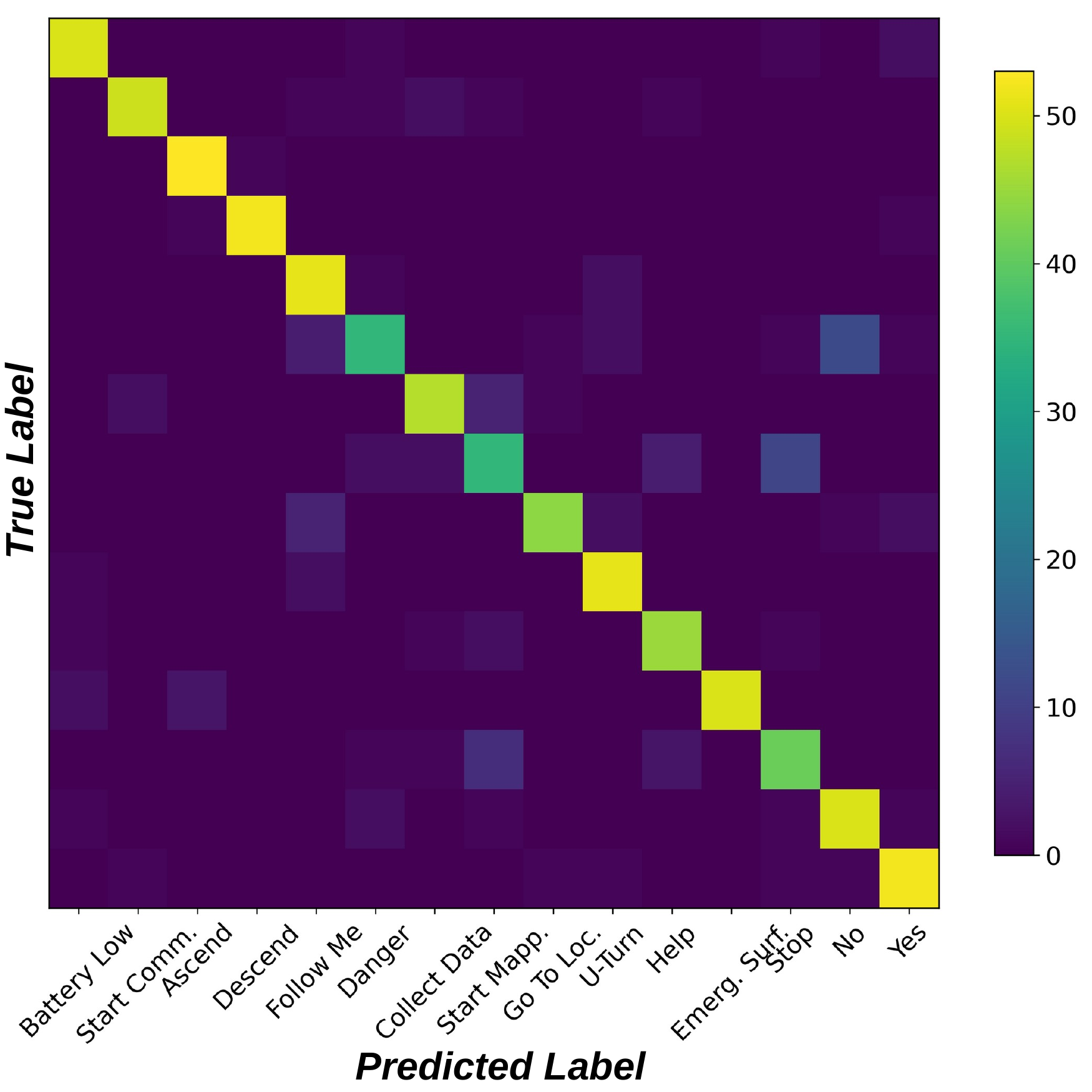}
    \caption{Confusion matrix for the gestural message recognition. It contains $5$ different experiments: RRCommNet in both simulated and real data, RRCommNet-Skip in both simulated and real data, and human transcription, performed on the $15$ communication messages.}
    \label{fig:cm}
    \vspace{-4mm}
\end{figure}

The test batch is fed to the trained RRCommNet model to get $10$ prediction scores as a tensor ($\textbf{x}_{preds}$) of shape $(10,15)$. We average the scores to get our final prediction vector, $\textbf{x}_{mean}$. Finally, we calculate the predicted class probability using the softmax function, as shown below:
\begin{equation}
    P(\textbf{x}_{mean}^i) = \frac{\exp (\textbf{x}_{mean}^i)}{\sum_{j=1}^N \exp (\textbf{x}_{mean}^j)} \label{eq:eval_softmax}
\end{equation}
where $i$ refers to the $i$-th class, $N$ is the total number of communication messages and $i=1,2,\dots,N$. Each prediction class is found by finding the maximum class probability and the respective index, $i$.

We consider three metrics for quantitative evaluation of RRCommNet: 
\begin{enumerate}
    \item \textbf{Recognition Accuracy:} it is the ratio between correct predictions and total instances (for each message).
    \item \textbf{Recognition Probability:} it is the softmax probability, as defined in~(\ref{eq:eval_softmax}), that shows the confidence for a correct prediction (for each message). 
    \item \textbf{Inference Time:} it is defined as the time it takes for each prediction (reported on CPU unless otherwise specified).
\end{enumerate}

\subsection{Results} \label{sec:results}
\subsubsection{RRCommNet Outperforms SOTA For AUV Gestures}
First, we compare the performance of the RRCommNets against the state-of-the-art (SOTA) action recognition models in terms of average recognition accuracy. From Table~\ref{tab:recog_acc}, we see that for simulated data, RRCommNet achieves an average recognition accuracy of $94.67$\% which is exactly same as the SOTA model, LateTemporal$3$DBert~\cite{kalfaoglu2020late}. In comparison, the SlowFast~\cite{feichtenhofer2019slowfast} model, which is another robust action recognition model, does not achieve comparable performance (accuracy $74.67$\%) against our method. As for real data, RRCommNet achieves an average recognition accuracy of $83.33$\% which is significantly better than the rest of the methods. 
In contrast, RRCommNet-Skip achieves an average recognition accuracy of $88\%$ on simulated data which is higher than the SlowFast method but only comparable against the SOTA or RRCommNet. For real data, however, RRCommNet-Skip shows superior performance than the SOTA.
As a note, we were unable to make a comparison with the underwater inter-robot communication framework presented in~\cite{koreitem2019underwater} because the authors in that paper use a 3D pose regressor as the visual decoder whereas we use a self-attention based classifier as our visual decoder.  

\subsubsection{RRCommNet Confusion Matches Human Confusion}
First, we evaluate the message recognition performance of both RRCommNet and RRCommNet-Skip by feeding our test data (described in Section \ref{sec:data}) to both networks and analyzing the outputs. 
Fig.~\ref{fig:vis} shows a snapshot of RRCommNet predicting the U-TURN message, performed by the Aqua AUV in actual underwater environment. 
The predictions are accumulated in a confusion matrix and further augmented by the human transcription of robotic conversation results (described in Sec.~\ref{sec:human_study}). 
Fig.~\ref{fig:cm} shows the complete confusion matrix where the values in different cells are normalized (refer to the color bar to see the non-normalized values). 
From the figure, we see that DANGER and START MAPPING are relatively difficult to recognize for humans and also by the RRCommNets. 
Analyzing the gesture implementations, we notice that DANGER and START MAPPING have a resemblance to NO and STOP, respectively. 
Noting this, we suggest that pre-deployment analysis of spatial similarity of gestures could optimize both robot and human comprehension of gestures by avoiding overlaps in spatio-temporal fragments.
\begin{table}[t]
\centering
\vspace{2mm}
\caption{Comparison of the gestural message recognition accuracy on both simulated and real data.}
\begin{tabular}{l|cc}
  \hline
  \begin{tabular}{@{}c@{}}\textbf{Method}\\\end{tabular} & \multicolumn{2}{c}{\textbf{Avg. Recognition Accuracy (\%)}} \\
  & \textbf{Simulated} & \textbf{Real} \\
  \hline \hline
  SlowFast &  $74.67$ &  $64.29$                    \\ \hline
  LateTemporal$3$DBert &  $94.67$ &      $71.42$   \\ \hline
 RRCommNet &  $94.67$ &   $83.33$    \\ \hline
  RRCommNet-Skip &  $88.00$ &    $73.81$   \\ \hline
\end{tabular}
\label{tab:recog_acc}
\vspace{-4mm}
\end{table}
\begin{table*}[t]
    \centering
    \vspace{2mm}
    \caption{Comparison between the performance of RRCommNet and RRCommNet-Skip in recognizing different gestural messages. The values represent \textit{avg. recog. probability (\%)$\Large/$avg. inference time (s)}.}
\begin{tabular}{l||cccccccccccccccc}
 \hline
\textbf{Gest. Message}  & BATTERY & START &  &  & FOLLOW &  & COLLECT & START \\
 
 & LOW & COMM. & ASCEND & DESCEND & ME & DANGER & DATA & MAPPING \\
 
\hline
RRCommNet  & $95.53/0.55$ & $78.16/3.48$ & $97.5/0.51$ & $96.94/0.59$ & $63.65/1.61$ & $47.93/2.37$ & $96.85/1.68$ & $65.00/0.51$ \\

RRCommNet-Skip  & $95.78/0.34$ & $80.93/2.37$ & $93.74/0.34$ & $98.16/0.35$ & $37.68/0.94$ & - & $98.48/1.01$ & $75.63/0.34$ \\
 
 &  &  &  &  &  &  &  &  \\
 
\hline
\textbf{Gest. Message}  & GO TO &  &  & EMERGENCY &  &  &  &  \\
 
   & LOCATION & U-TURN & HELP & SURFACING & STOP & NO & YES & \textbf{Overall} \\ 
 
 \hline
RRCommNet   & $86.68/1.67$ & $64.06/0.75$ & $59.5/4.72$ & $98.47/0.73$ & $66.9/1.32$ & $73.1/0.67$ & $94.3/0.66$ & $78.97/1.45$ \\

RRCommNet-Skip  & $78.81/1.03$ & $55.61/0.55$ & $68.49/3.07$ & $98.69/0.35$ & $61.08/0.90$ & $63.14/0.41$ & $98.07/0.47$ & $78.88/0.80$ \\
\hline

\end{tabular}
\label{tab:infer}
\vspace{-4mm}
\end{table*}

\subsubsection{RRCommNet-Skip Outperforms RRCommNet in Speed}
Finally, we evaluate the performance of the RRCommNets in terms of their prediction confidence and speed. From Table~\ref{tab:infer}, we see that both RRCommNets are fairly confident while making gestural message predictions, having overall average recognition probability of $78.97$\% and $78.88$\%, respectively. As for recognition speed, we see that RRCommNet-Skip is notably faster than RRCommNet. 
As a matter of fact, both the networks display fast inference times on CPU. 
For example, inference times for BATTERY LOW and STOP messages are $0.55$s and $1.32$s, respectively using RRCommNet and $0.34$s and $0.90$s, respectively using RRCommNet-Skip. Therefore, we can choose RRCommNet-Skip where inference time is important, and choose RRCommNet where accuracy is the main requirement.

\subsection{Runtime Performance on Embedded GPU}
As established in the previous section, there is a trade-off between inference time and accuracy with RRCommNet and RRCommNet-Skip, as shown in Table \ref{tab:infer}.
However, these inference times are on a powerful CPU (an Intel\textregistered{} Xeon\textregistered{} E5-2650), processing power that most AUVs will not have available onboard. 
It is, therefore, worthwhile to consider runtime performance on a processor more likely to be used in an AUV, such as the Nvidia TX2 (already in use in a number of AUVs, Aqua among them).
To test on the TX2, we simply run our evaluation experiments a second time, recording the inference time for each video, calculating the per-frame inference time for both networks, and averaging across all videos.
RRCommNet operates at an average inference speed of \textbf{10 FPS} and RRCommNet-Skip performs inference at \textbf{12.5 FPS}.
Total inference time for any gesture depends on the length of the gesture, but this rate of inference allows the recognition of gestures at sufficient speeds to be useful in the field.

\section{CONCLUSION}
In this work, we have shown that an attention mechanism-based deep network can recognize a gestural language for AUVs in both simulated and real environments.
Our networks, RRCommNet and RRCommNet-Skip, demonstrated recognition accuracy which outperforms state-of-the-art activity recognition methods.
Both networks also have inference speeds on embedded GPU which are sufficient for onboard robot deployments, though RRCommNet-Skip is somewhat faster.
Through a human transcription study, we have also demonstrated that the proposed gestural language is understandable to humans. 
Having shown our framework to be comprehensible to both humans and robots, we also suggest that our framework is ideal for integration into existing diver interaction workflows. 
The majority of underwater communication between divers happens via gestures (with some exceptions). 
In much the same way, human-to-robot control methods underwater often utilize gestures~\cite{islam2018dynamic,chavez_2021_gesture}. 
Thus, a gesture-based robot-to-robot language fits nicely into the overall workflow of underwater work. 
On the technical side, our ROS-based implementation makes the process of integrating our method into any ROS-powered AUV relatively straightforward.
Lastly, implementation of the gestures in our language via simple configuration files makes the transition from designing a gesture to implementing it as simple as writing out the desired movements. 
While programming experience will still be required to fine-tune the movements, this implementation will allow for the easy creation of new gestures when they are required.
Taken together, our method's state-of-the-art recognition performance, reasonable inference times, human comprehensibility, conformity with existing robot and human underwater communication methods, and easy generation/reconfiguration of gestures make it an excellent candidate for deployment in the field.
This enables the opportunity to use our gestural language and language recognition network for AUV-to-AUV communication in underwater m/HRI missions, so that every party to the interaction, whether human or robot, can understand every part of the conversation.


\bibliographystyle{ACM-Reference-Format}
\bibliography{root}


\end{document}